\documentclass[sigconf,screen]{acmart}

\AtBeginDocument{%
  }

\setcopyright{acmcopyright}
\copyrightyear{2023}
\acmYear{2023}
\acmDOI{XXXXXXX.XXXXXXX}

\acmConference[ASPLOS’23]{Make sure to enter the correct
  conference title from your rights confirmation emai}{March 25--29,
  2023}{Vancouver, Canada}
\acmPrice{15.00}
\acmISBN{978-1-4503-XXXX-X/18/06}

\usepackage{xspace}
\usepackage{subfigure}
\usepackage[font=footnotesize,labelfont=bf]{caption}
\usepackage[normalem]{ulem}
\usepackage{amsmath}
\usepackage{multirow}

\newcommand\ie{\emph{i.e.,\ }}

\newcommand{\id}[0]{\textsf{id}\xspace}
\newcommand{\num}[0]{\textsf{num}\xspace}
\newcommand{\SPTypeSet}[0]{\mathbf{T}\xspace}

\begin{document}

\title{TLP: A Deep Learning-based Cost Model \\
for Tensor Program Tuning}

\author{Yi Zhai}

\affiliation{%
  \institution{University of Science and Technology of China}
  \city{Hefei}
  \country{China}
}
\email{zhaiyi0@mail.ustc.edu.cn}

\author{Yu Zhang}
\authornote{Corresponding author.}
\affiliation{%
  \institution{University of Science and Technology of China}
  \city{Hefei}
  \country{China}}
\email{yuzhang@ustc.edu.cn}

\author{Shuo Liu}
\affiliation{%
  \institution{University of Science and Technology of China}
  \city{Hefei}
  \country{China}}
\email{zkdliushuo@mail.ustc.edu.cn}

\author{Xiaomeng Chu}
\affiliation{%
  \institution{University of Science and Technology of China}
  \city{Hefei}
  \country{China}}
\email{cxmeng@mail.ustc.edu.cn}

\author{Jie Peng}
\affiliation{%
  \institution{University of Science and Technology of China}
  \city{Hefei}
  \country{China}}
\email{pengjieb@mail.ustc.edu.cn}

\author{Jianmin Ji}
\affiliation{%
  \institution{University of Science and Technology of China}
  \city{Hefei}
  \country{China}}
\email{jianmin@ustc.edu.cn}

\author{Yanyong Zhang}
\affiliation{%
  \institution{University of Science and Technology of China}
  \city{Hefei}
  \country{China}}
\email{yanyongz@ustc.edu.cn}

\renewcommand{\shortauthors}{}

\begin{abstract}

Tensor program tuning is a non-convex objective optimization problem, to which search-based approaches have proven to be effective. 
At the core of the search-based approaches lies the design of the cost model. Though deep learning-based cost models perform significantly better than other methods, they still fall short and suffer from the following problems. First, their feature extraction heavily relies on expert-level domain knowledge in hardware architectures. Even so, the extracted features are often unsatisfactory and require separate considerations for CPUs and GPUs. Second, a cost model trained on one hardware platform usually performs poorly on another, a problem we call cross-hardware unavailability.

In order to address these problems, we propose TLP and MTL-TLP. TLP is a deep learning-based cost model that facilitates tensor program tuning. Instead of extracting features from the tensor program itself, TLP extracts features from the schedule primitives. We treat schedule primitives as tensor languages. TLP is thus a \textbf{T}ensor \textbf{L}anguage \textbf{P}rocessing task. In this way, the task of predicting the tensor program latency through the cost model is transformed into a natural language processing (NLP) regression task. MTL-TLP combines \textbf{M}ulti-\textbf{T}ask \textbf{L}earning and \textbf{TLP} to cope with the cross-hardware unavailability problem.

We incorporate these techniques into the Ansor framework and conduct detailed experiments.
Results show that TLP can speed up the average search time by 9.1$\times$ and 3.0$\times$ on CPU and GPU workloads, respectively, compared to the state-of-the-art implementation. MTL-TLP can achieve a speed-up of 4.7$\times$ and 2.9$\times$ on CPU and GPU workloads, respectively, using only 7\% of the target hardware data.
To the best of our knowledge, TLP is the first tensor program cost model to extract features directly from schedule primitives, and MTL-TLP is the first open-sourced work that effectively addresses the cross-platform unavailability problem.
The code is available at \href{https://github.com/zhaiyi000/tlp}{https://github.com/zhaiyi000/tlp}.

\end{abstract}



\received{7 July 2023}
\received[revised]{27 October 2023}
\received[accepted]{3 November 2023}

\maketitle


\section{Introduction}

The main challenge of tensor program tuning is that measuring tensor program latency is excessively time-consuming. 
The time-consuming measurement of the tensor program latency is due to the following three reasons. First, the measurement pipeline consists of multiple steps, including compilation, loading, and execution. Two network transfers are required if the remote procedure call method is used. The compilation is particularly time-consuming, requiring lowering to multiple intermediate representations (IRs) and performing extensive optimizations on these IRs. 
Second, to guarantee measurement accuracy, it is often necessary to repeat the measurement multiple times. If the measurement is performed on CPUs, the cache needs to be flushed between two consecutive measurements. Finally, the measurement task often monopolizes the computing resources to avoid noises, preventing the potential parallel execution mode. In our experience, measuring the latency of a tensor program for a fusion operator often takes hundreds of milliseconds (on Intel i7-10510U CPU and NVIDIA GeForce 2080Ti).

The time-consuming problem makes it impossible to place all the generated tensor programs on the target hardware platform for latency measurement during the tuning process. To address this issue, many deep learning compilers~\cite{baghdadi2019tiramisu, chen2018tvm, XLA} resort to the cost model~\cite{baghdadi2021deep, chen2018learning, kaufman2021learned, zheng2020ansor, zheng2021tenset} --  using the predicted latency from the cost model as the criterion. Dozens of candidates are selected to be measured on target hardware from tens of thousands of tensor programs. Recent works~\cite{baghdadi2021deep, steiner2021value, zheng2021tenset} have demonstrated that deep learning-based cost models perform far better than other methods. However, existing deep learning-based cost models rely heavily on complex feature engineering. Ansor~\cite{zheng2020ansor} manually extracts 164 features from five aspects for the innermost assignment statement, including computation, memory access, and arithmetic strength. The TIRAMISU cost model~\cite{baghdadi2021deep} manually extracts 2534 features and the abstract syntax tree (AST) structure; it takes the AST structure as a computational stream to forward propagate input features. Complex feature engineering can be problematic. First of all, only domain experts well-versed in hardware architecture can be qualified for the job. Even so, the hand-picked cost models still fall short in many cases, largely affected by  the limitation of prior knowledge. More importantly, the features extracted on CPUs and GPUs are different; even when extracted on CPUs, the features cannot achieve stable performance simultaneously on different architecture CPUs such as Intel, ARM, and AMD CPUs. In order to take into consideration the different hardware architectures, more engineering endeavors are required, making feature extraction a laborious and cumbersome process. 

This time-consuming problem also hinders us from collecting large-scale tensor program datasets on every possible hardware platform. Furthermore, the performance of a cost model trained offline on one hardware platform drops dramatically on another, which we call the problem of offline cost model unavailability across hardware. A tensor program has different execution times on different hardware, sometimes with a considerable variance  due to domain gaps in hardware architecture and hardware performance. Such discrepancy makes it hard for a model to apply simultaneously on multiple hardware platforms.  To address this issue, TenSet~\cite{zheng2021tenset} and Moses~\cite{zhao2022moses} use transfer learning and model distillation, respectively.

In response to these problems, we present TLP and MTL-TLP. TLP is a deep learning-based cost model for tensor program tuning. We do not extract features from the tensor program source code for two reasons. First, the source code of the tensor program is tree-structured data with nested loops, and the AST information in it is difficult to extract and utilize. The second is that there are too many unrelated character tokens in the source code, resulting in sparse features. We extract features from schedule primitives. We treat schedule primitives as tensor language and turn the task of predicting tensor program latency into an NLP regression task.

For the problem of the offline cost model being unavailable across hardware, we use multi-task learning techniques to solve it. Multi-task learning has led to successes in many deep learning applications, from language translation~\cite{dong2015multi, niehues2017exploiting} and text classification~\cite{liu2016recurrent} to scene analysis~\cite{kendall2018multi} and autonomous driving~\cite{chen2018multi, chowdhuri2019multinet}. Similar to the idea that the tensor compiler uses high-level graph IR and low-level tensor IR to abstract hardware-independent and hardware-dependent features, we use shared and non-shared parameters to fit hardware-independent and hardware-dependent features. One tensor program has different latencies (labels) on different hardware platforms. We set up multiple tasks, each corresponding to a hardware platform.  

We integrate the above techniques into Ansor~\cite{zheng2020ansor}, a state-of-the-art search framework in the TVM compiler.
We conduct comprehensive experiments and show that TLP and MTL-TLP comprehensively outperform the state-of-the-art implementation on various CPU and GPU workloads.
Compared with Ansor, reaching the performance of Ansotr tuning 2,000 times, TLP can speed up the average search time by 16.7$\times$ and 16.0$\times$ on CPU and GPU workloads. MTL-TLP can achieve a speed-up of 10.0$\times$ and 15.8$\times$ on CPU and GPU workloads, respectively. In many experiments, there is still a big gap between the performance of Ansor tuning 2000 times and the performance of TLP/MTL-TLP tuning the minimum times. The minimum times refers to the total times of tuning one round for each subgraph of the workload only. If Ansor is given more tuning time budgets, this speed-up can be much larger. Compared with TenSet, TLP can speed up the average search time by 9.1$\times$ and 3.0$\times$ on CPU and GPU workloads, reaching the performance of TenSet MLP tuning 2,000 times. MTL-TLP can speed up the search time by 4.7$\times$ and 2.9$\times$ on CPU and GPU workloads, respectively. Among them, MTL-TLP only uses 7\% of the target hardware platform data.

It is worth noting that TLP and MTL-TLP are not specifically designed for Ansor. TLP and MTL-TLP apply to all automatic search frameworks that lower high-level graph IR into low-level tensor IR via schedule primitives. 

In summary, this paper makes the following contributions:
\begin{itemize}
    \item A simple yet efficient and general feature extraction mechanism for tensor programs;
    \item A multi-task learning method to address offline cost model unavailability across hardware platforms;
    \item An implementation and comprehensive evaluation of the TLP and MTL-TLP demonstrate that the above techniques outperform state-of-the-art implementation on various deep learning models and hardware platforms.
\end{itemize}

\section{Background}

\begin{figure}[t]%
    \centering
    \includegraphics[scale=0.7]{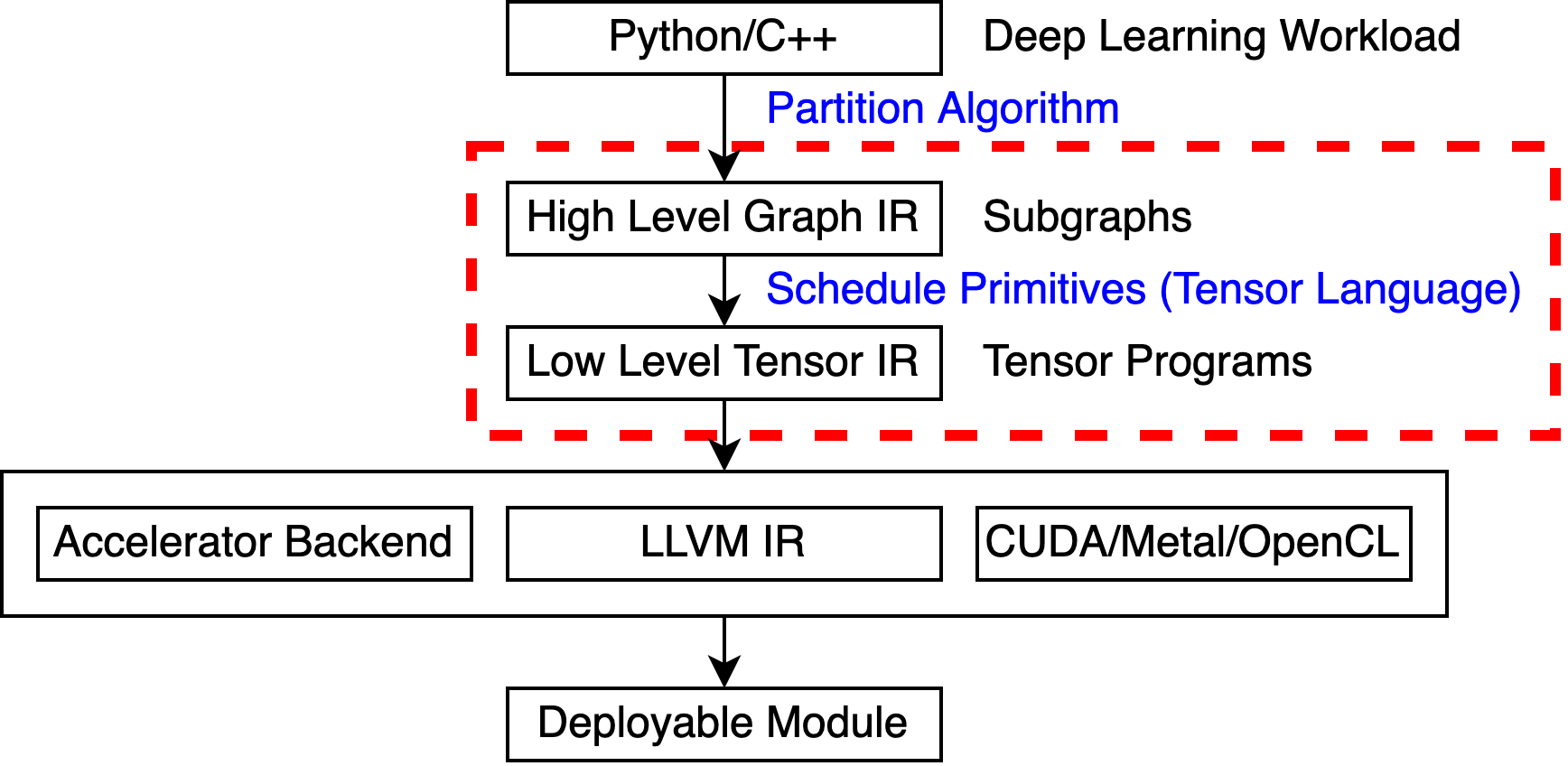}  

    \caption{The pipeline of common deep learning compilers. The red dashed box is the workspace of TLP.}%
    \label{2_1_compiler_pipeline}%

\end{figure}
\textbf{Search-based compilers.} Figure \ref{2_1_compiler_pipeline} shows the pipeline of common deep learning compilers. These compilers accept deep learning workloads as input and output executables for specific hardware platforms. A deep learning workload is essentially a computational graph written in various high-level programming languages. After several computational graph optimizations, some partitioning algorithms divide the long computational graph into computational subgraphs. Each subgraph has its own search space, determined by the input/output tensor shapes, data types, data layouts, and the target hardware platform. The search space is usually in the order of millions on CPUs and billions on GPUs. Guided by genetic algorithms (GA), Model Carlo tree search (MCTS), simulated annealing, etc., the compiler applies various combinations of schedule primitives to each subgraph. In the process, tens of thousands of tensor programs are generated, and the compiler uses a cost model to filter out candidates with the best possible performance. Then put these candidates on the target hardware to measure to find the highest performing tensor program. These high-performance tensor programs are finally delegated to specific accelerated backends or well-developed LLVM/CUDA to generate the final executable. 
\begin{figure*}[t]%
    \centering
    \includegraphics[scale=0.6]{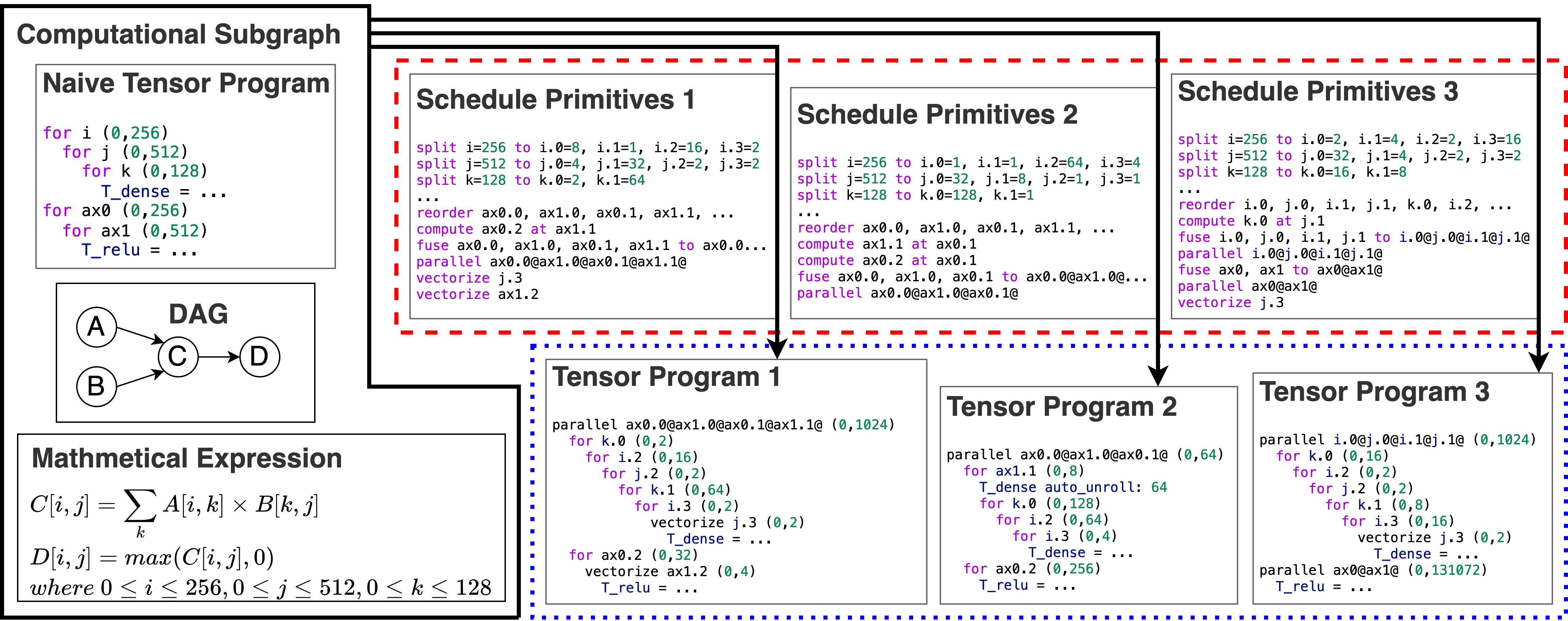}  

    \caption{Subgraph, a fused dense + ReLU activation, applies combinations of different schedule primitives to generate tensor programs. The mathematical expression, DAG, and naive tensor program contained in the computational subgraph above are logically equivalent but in different forms. Schedule primitives are written in pseudocode. The schedule primitives in the red long dashed box are the feature extraction objects of TLP, and the tensor programs in the blue short dashed box are the feature extraction objects of Ansor~\cite{zheng2020ansor}, TIRAMISU cost model~\cite{baghdadi2021deep}, etc.}%
    \label{2_2_relay_to_tensor}%

\end{figure*}
Figure \ref{2_2_relay_to_tensor} takes a computational subgraph as an example to expand the red dashed box in Figure \ref{2_1_compiler_pipeline}. The three tensor programs are computationally equivalent, only with different latencies.

\textbf{TenSet.} 
TenSet~\cite{zheng2021tenset} proposes a large-scale tensor program dataset and builds a multilayer perceptron (MLP) tensor program cost model pre-trained with the dataset. Later, we use the TenSet dataset and TenSet MLP to represent its dataset and cost model, respectively.
The TenSet dataset was collected on 6 hardware platforms, including Intel CPUs, AMD CPUs, ARM CPUs, and NVIDIA GPUs. On each hardware platform, 2,308 subgraphs divided from 120 typical deep learning workloads such as ResNet, MobileNet, and BERT are collected, a total of $2,308\times6=13,848$ subgraphs on the 6 hardware platforms. For each subgraph, at most 4,000 tensor programs are generated by the automatic search framework Ansor integrated in TVM. In total, this dataset includes approximately 51.57 million tensor programs and their latency. Each hardware platform has approximately 8.59 million data. TenSet is a continuation of Ansor. Compared to Ansor, the TenSet MLP can speed up the search time by up to 10$\times$. We use Ansor and the TenSet MLP as the baselines for TLP and MTL-TLP.

We extensively analyze the TenSet dataset and use the dataset to train and evaluate TLP and MTL-TLP. At the same time, we also spent more than 50 days collecting a TenSet CPU dataset on an Intel i7-10510U CPU, which contains approximately 8.65 million data. In the following, we call this dataset the TenSet-TLP dataset.  

\textbf{Natural language processing.}
As a branch of artificial intelligence technology, natural language processing (NLP) uses deep learning to process and interpret text data. It has achieved great success in sentiment analysis, chatbots, machine translation, and more. The processing objects of an NLP task are usually sentences, ordered sequences composed of words. For many NLP tasks, they first project each word of sentences to a corresponding number called a token, then convert each token into a high-dimensional vector (denoted as embedding vector), and this convert operation is usually called embedding. A suitable embedding method can make certain distances between synonyms, such as Euclidean distance and cosine distance, short.

\section{System Overview}

\begin{figure}[t]%
    \centering
    \includegraphics[scale=0.5]{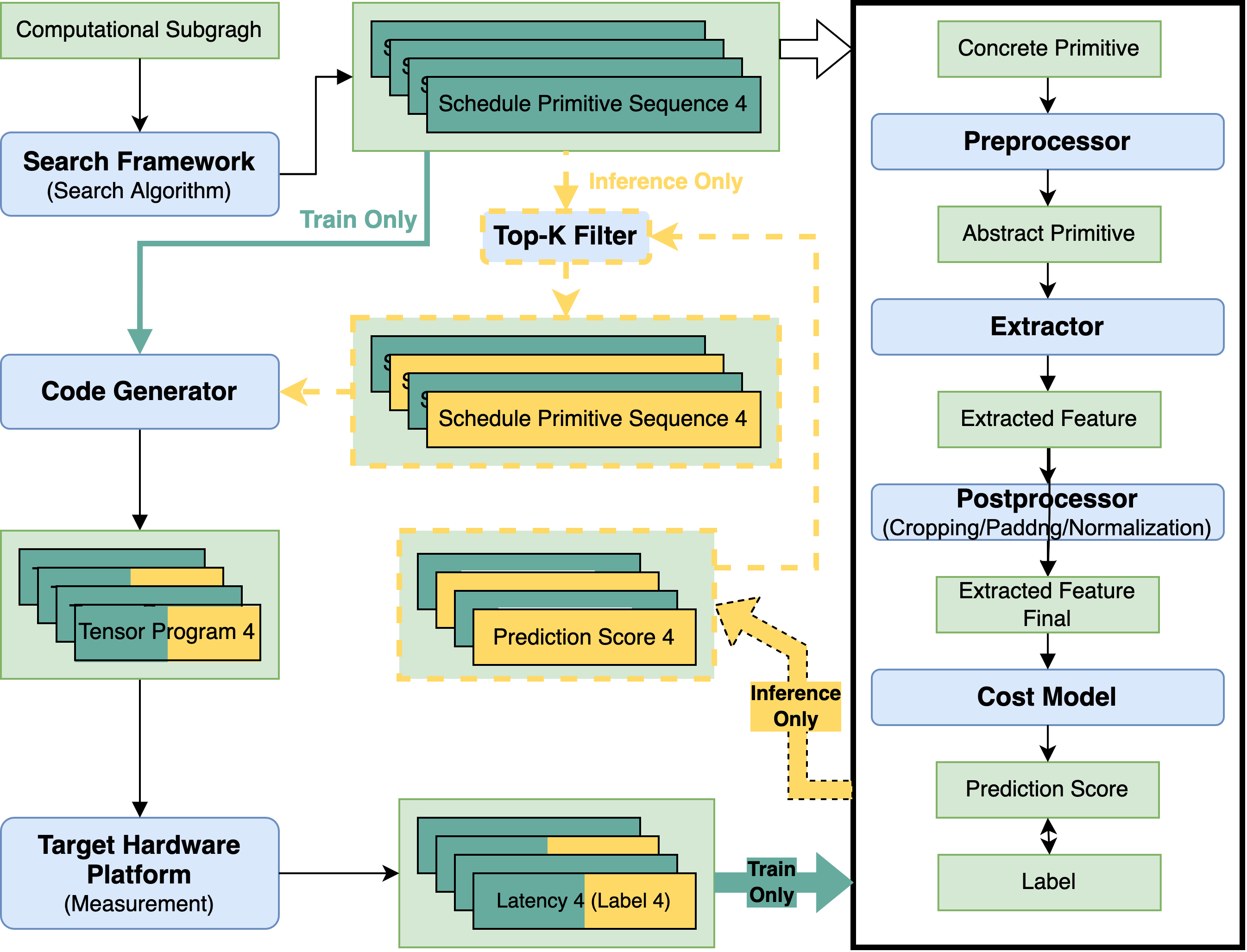}  

    \caption{The process of generating tensor programs from computational subgraphs by an automatic search framework integrated with TLP. The black box on the right is the pipeline of TLP. Tensor programs and latency (label) blocks with dark green on the left and yellow on the right are seen as dark green during training and yellow during inference. When training, the dashed arrows, dashed boxes, and "Inference Only" arrows in the figure are invalid. When inference, the "Train Only" arrows in the figure are invalid.}%
    \label{3_1_system_overview}%

\end{figure}
Figure \ref{3_1_system_overview} shows the process of generating tensor programs from computational subgraphs by an automatic search framework (abbreviated as auto-tuner) integrated with TLP. The black box on the right is the pipeline of TLP. The figure depicts the two states of TLP training and inference. When training, the auto-tuner generates several schedule primitive sequences according to the received computation subgraph. The auto-tuner then invokes the code generator to generate tensor programs in combination with the computational subgraph and the schedule primitive sequence and measures the latency of these tensor programs on the target hardware. TLP takes the schedule primitive sequence as input. After preprocessing, feature extraction, and post-processing, the TLP cost model forwardly propagates the finally extracted features and normalizes the latency of the corresponding tensor program as a label to calculate the loss. Finally, the loss is back-propagated to update the weights of the cost model. When inference, after generating the schedule primitive sequences, the auto-tuner obtains the prediction score through the cost model and screens out the top-k potential candidates (yellow blocks in the figure) according to the prediction score. The auto-tuner finally generates tensor programs for these candidates and measures their latency on the target hardware.

In the pipeline of TLP, the concrete primitive refers to the schedule primitive in an automatic search framework. The abstract primitive refers to the primitive conforming to the TLP feature extractor's input specification after preprocessing. We describe the preprocessor, extractor, and postprocessor in detail in Section \ref{TLP}.

\section{TLP}\label{TLP}
Feature extraction is the core of building a cost model. It determines the upper bound of the performance that the cost model can achieve. Ansor manually extracts 164 features from five aspects for the innermost assignment statement, including computation, memory access, and arithmetic strength. TenSet MLP follows Ansor's feature extraction and adds 10 high-level computational graph IR features. The TIRAMISU cost model is also extracting features from the innermost assignment statement. It extracts 2534 features, but most are zero, so its features are sparse. Unlike Ansor, the TIRAMISU cost model extracts the AST structure as a computational stream to forward propagate input features. However, in our opinion, treating the AST structure as a computational stream is an impracticable solution. The AST of each tensor program is different, so that the cost model can only forward propagate one by one without parallelism. The TIRAMISU cost model can only run on CPUs. Other pitfalls of these methods have been discussed in previous chapters.

\subsection{Feature Extraction of TLP}
To the best of our knowledge, TLP is the first tensor program cost model to extract features directly from schedule primitives. It is challenging to efficiently extract features from schedule primitives and retain all valuable information significantly. There are three methods here. Method 1: Treat the schedule primitive sequence as character text and then use NLP tasks to do text processing. Method 2: Treat each schedule primitive as an NLP word, so a complete  schedule primitive sequence is an NLP sentence. Then, convert each word into a token. Method 3, adopted by TLP, regards schedule primitive as a combination of three basic elements, \ie primitive type, numeric parameters, and character parameters. Extract features for the three elements separately and then concatenate the extracted features together. This paper does not discuss the first method, for this simple and crude end-to-end method is challenging to achieve good performance. We think the second method misses some helpful information on the table. For example, primitives of the same type but different parameters will be encoded into two tokens without any relationship. Only the embedding algorithm and deep learning network can mine the synonymous relationship between them. Method 3 powerfully preserves this synonym relationship. The detailed description is as follows.

\begin{figure}[t]
\begin{scriptsize}
\subfigure[Abstract Schedule Primitive Sequence]{
$\begin{array}{lrcl}
\text{(PrimitiveSequence)} &
S    &::=& p^{*}  \\
\text{(Primitive)} &
p    &::=& \tau~(\id ~|~ \num)^{*} \\
\text{(PrimitiveType)} &
\SPTypeSet\ni \tau &::=& \text{split | reorder | fuse | ...} \\
\text{(NameParam)} &
\id    &&\\
\text{(Number)} &
\num    &&\\
\end{array}$
}
~\\
\subfigure[TLP Extractor]{
$\begin{array}{lrcl}
\text{(Features)} &
f=F(p)&::=&F_{1}(\tau)\left (  F_{2}{ \left( \id \right )}|F_{3}(\num) \right )^{*} \\

&\multicolumn{3}{l}{F: \text{Primitive} \rightarrow \text{Features}}\\
&\multicolumn{3}{l}{F_{1}: \text{PrimitiveType}\rightarrow \text{OnehotVector}}\\
&\multicolumn{3}{l}{F_{2} : \text{NameParam} \rightarrow \text{Token}}\\
&\multicolumn{3}{l}{F_{3} : \text{Number} \rightarrow \text{Number}}\\
\end{array}$
}
\end{scriptsize}
\caption{The specification of TLP feature extraction.}
\label{3_1_feature_extraction_abstract}
\end{figure}
\begin{figure}[t]%
    \centering
    \includegraphics[scale=0.6]{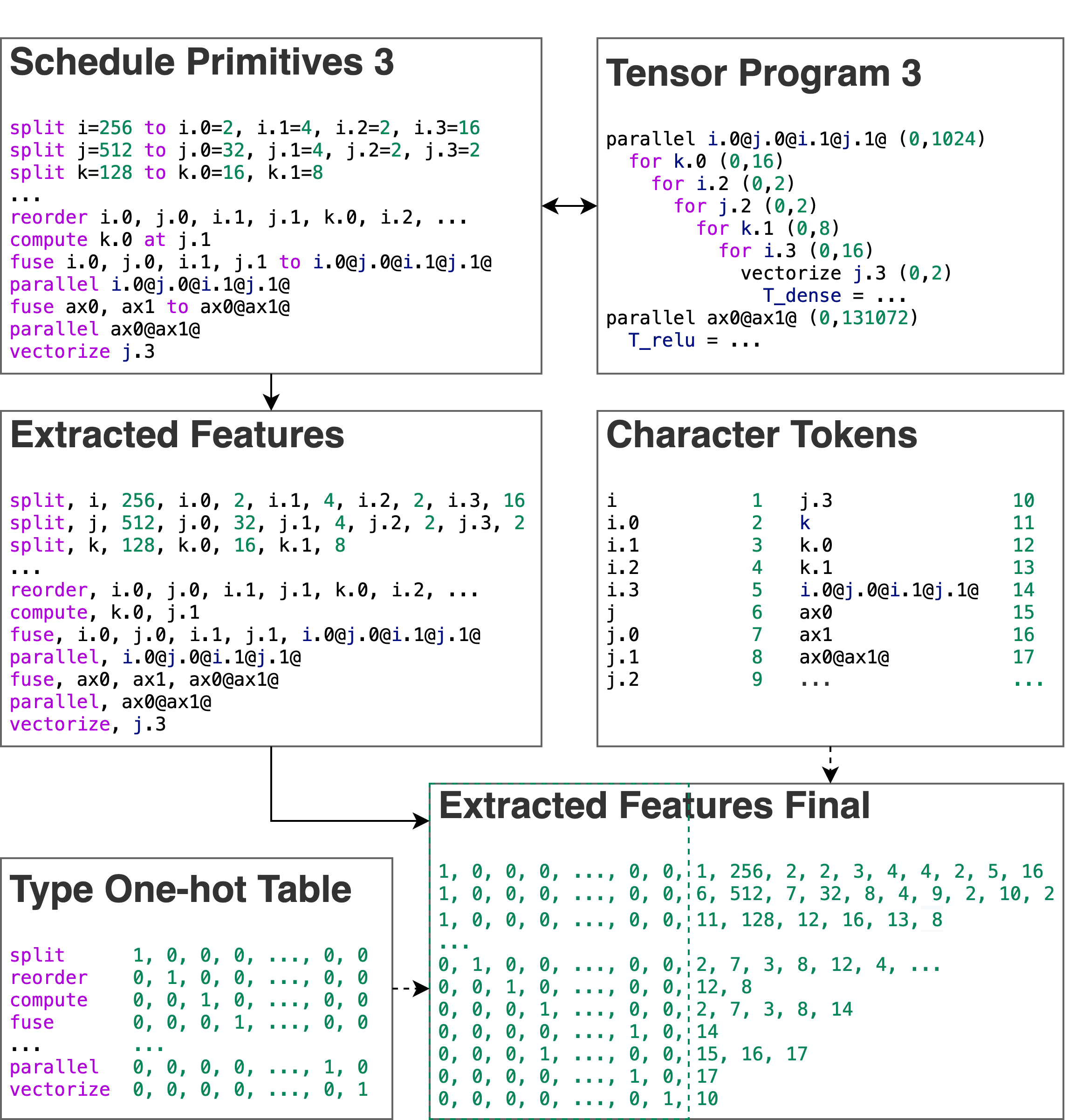}  

    \caption{A feature extraction example for TLP.}%
    \label{3_1_feature_extraction}%

\end{figure}
TLP feature extraction first preprocesses the input schedule primitive sequence. For each primitive in the sequence, TLP feature extraction only retains three basic elements: primitive type, numeric parameters, and character parameters. Extraneous characters will be stripped. In fact, these three elements basically contain all the semantic information in the schedule primitive. The preprocessing algorithm implementation is related to the specific automatic search framework. In most frameworks, this preprocessing algorithm is reversible, i.e., the schedule primitive sequence can be recovered from these three basic elements since each primitive has its semantics and input specification. Figure \ref{3_1_feature_extraction_abstract} is the specification of TLP feature extraction. Figure \ref{3_1_feature_extraction_abstract}a depicts an abstract representation of schedule primitives. A schedule primitive sequence (PrimitiveSequence) consists of several schedule primitives (Primitive). Each primitive starts with the primitive type (PrimitiveType), followed by several numeric parameters (Number) and character parameters (NameParam). The primitive type is determined by the specific automatic search framework. Figure \ref{3_1_feature_extraction_abstract}b is an implementation of the extractor for TLP feature extraction. For primitive types, convert it to a one-hot vector. The one-hot vector is determined by the specific automatic search framework. For numeric parameters, leave their value unchanged. For character parameters, convert them into tokens, the same way NLP tasks deal with words. We map different character parameters to different tokens. Finally, all the features are concatenated according to the element's original position. After that, the extracted features are post-processed by methods such as cropping, padding, and normalization. 

We take Figure \ref{3_1_feature_extraction} as an example to illustrate the process of TLP feature extraction. The tensor program 3 and schedule primitive 3 in Figure \ref{3_1_feature_extraction} are extracted from Figure \ref{2_2_relay_to_tensor}. First of all, it is clear that the task of TLP is to estimate the latency of the tensor program, that is, to estimate the latency of tensor program 3 in Figure \ref{3_1_feature_extraction}. We regard tensor program 3 and schedule primitive 3 as equivalents, and the theoretical basis will be analyzed later. Figure \ref{3_1_feature_extraction} shows an example of TLP feature extraction from top to bottom, \ie the Schedule primitives 3, the Extracted Features, and the Extracted Features Final. Figure \ref{3_1_feature_extraction} has no post-processing process.

We can think of a primitive as an NLP word and the encoded result as the embedding vector of the NLP task after going through the embedding layer. The TLP feature extraction is a word embedding algorithm, which significantly preserves the synonym relationship. Since the encoding results of two primitives of the same type but different parameters have many of the same values, the Euclidean distance between the two primitives after normalization is relatively short. In addition, TLP directly extracts features from schedule primitives, eliminating the need to generate tensor programs and reducing tuning time.

\subsection{TLP Feature Extraction on TenSet dataset}
\textbf{Terminology.} We define some terminologies used in the rest of this paper. Sequence length: the length of a schedule primitive sequence. Embedding size: the number of features extracted from a schedule primitive. Feature size: the number of all features extracted from a schedule primitive sequence for a tensor program, \ie sequence length $\times$ embedding size.

Let us take a look at the TLP feature extraction on the TenSet dataset to gain an intuitive understanding.
\begin{table}[b]
\scriptsize
\centering
\begin{tabular}{|c|c|c|c|c|c|c|c|}
\hline
RE & 40 & FU & 22 & SP & 18 & FSP & 15 \\ \hline
CA & 14 & AN & 14 & RF & 14 & PR & 14 \\ \hline
CHW & 13 & CR & 12 & CI & 12 & & \\ \hline
\end{tabular}
\caption{The maximum embedding sizes of each schedule primitive in the TenSet CPU dataset. “RE”, “FU”, etc. in the table are the abbreviations of schedule primitives, which refer to primitives such as reorder and fuse, respectively.}
\label{3_1_step_len}
\end{table}
\begin{figure}[t]%
    \centering
    
    \includegraphics[scale=0.6]{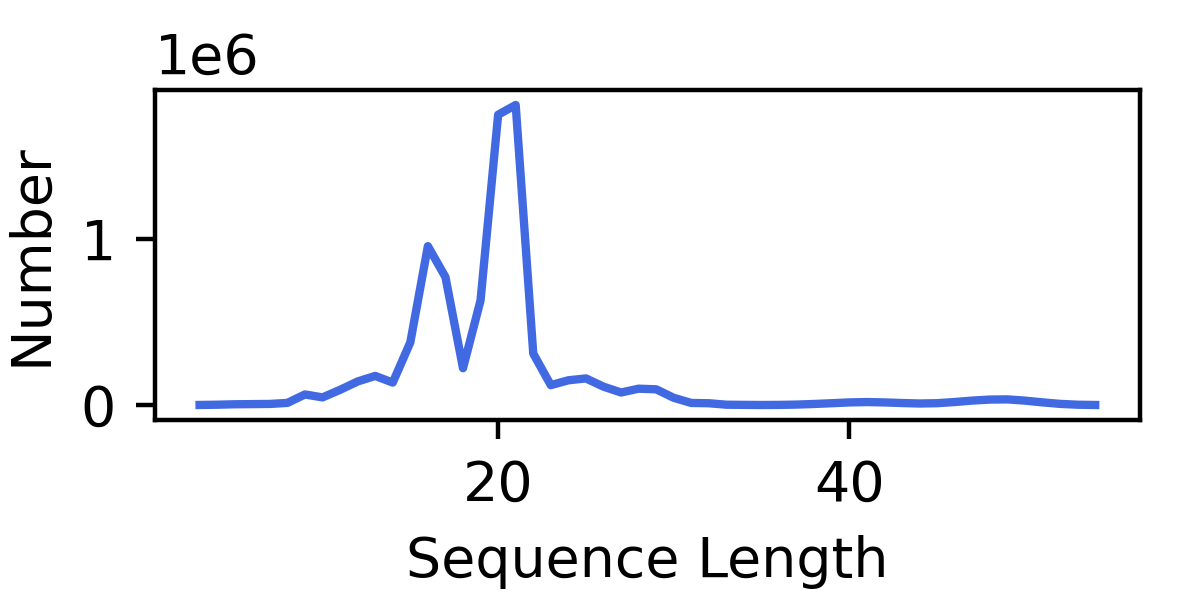}  

    \caption{The distribution of sequence lengths for tensor programs in the TenSet CPU dataset.}%
    \label{3_2_line_len}%

\end{figure}
TenSet uses the automatic search framework Ansor to dump the tensor programs. Ansor uses 11 types of schedule primitives in CPUs and GPUs, respectively, so the types of all schedule primitives are replaced by an 11-dimensional one-hot vector. According to our statistics, in the TenSet CPU dataset, the schedule primitive sequence of tensor programs contains up to 54 primitives, each primitive type plus its parameters up to 40 (after the type is changed to an 11-bit one-hot vector). Figure \ref{3_2_line_len} shows the distribution of sequence lengths of tensor programs in the TenSet CPU dataset. Of these, at most 1,807,960 tensor programs have schedule primitive sequences of the same length 21. The embedding size for a primitive is indeterminate. For example, a split primitive can split a loop variable to 3 or 4 loop variables. Table \ref{3_1_step_len} shows the maximum embedding sizes of each schedule primitive in the TenSet CPU dataset. These data suggest that even limiting the feature size of a tensor program from $54\times40=2,160$ to $25\times22=550$ would not hurt the vast majority of features. Subsequent analysis and experiment also prove this. The statistics of the TenSet GPU dataset are similar and will not be repeated here.

\subsection{Feasibility Analysis of TLP Feature Extraction} 
Are schedule primitive sequences definitely different for different tensor programs? The answer to this is no. Nevertheless, we think the probability of different tensor programs having the same  schedule primitive sequence is low. The same computational subgraph combined with the same  schedule primitive sequence must generate the same tensor program. For different computational subgraphs, either their computational flow is different, or their computational parameters are different. We think that a high-performance schedule primitive sequence contains most or even all computational parameters of a computational subgraph, and the schedule primitive sequence is tailored according to the computational flow of the computational subgraph. This results in a low collision probability of different computation subgraphs but the same schedule primitive sequence. Taking a step back, even if this collision probability rises, the feature extraction scheme of TLP is still available. The optimal value of the labels of multiple tensor programs corresponding to a schedule primitive sequence can be used as the label of the schedule primitive sequence. The disadvantage of this approach is that it may cause the cost model to misjudge a low-performance tensor program as a high-performance tensor program, resulting in a longer tuning time. Nevertheless, it guarantees that a high-performance schedule primitive sequence will not be missed.

We analyze the TenSet CPU database and find that there are 8.56 million different schedule primitive sequences in a total of 8.65 million tensor programs. The repetition rate is only 1.0430\%. Even if the features are limited to 25$\times$22, there are 8.53 million differences out of 8.56 million tensor programs, with a repetition rate of 1.4034\%. Therefore, we can approximately think that a schedule primitive sequence uniquely characterizes a tensor program. That is, we consider a schedule primitive sequence equivalent to a tensor program. 

The advantages of schedule primitive sequences are apparent. First, as mentioned above, in the TenSet CPU dataset, a tensor program can be characterized almost entirely by 550 features. These features are dense and preserve all information considerably. Second, the schedule primitive sequence is a regular sequence that can take advantage of some existing NLP techniques. Third, there are very few kinds of schedule primitives, such as only 14 (including for CPUs and GPUs) in Ansor, and most of them are common to both CPUs and GPUs. Finally, For schedule primitives, we do not need complex feature engineering. We just record the type and parameters of schedule primitives. It is possible to avoid operating on complex nested loops in the AST.

As such, the schedule primitive sequence and the tensor program have an almost one-to-one correspondence, containing the same amount of semantic information. The source code of the tensor program is a nested loop tree structure, and the  schedule primitive sequence is a regular sequence structure. From the perspective of language, the structure of a schedule primitive sequence is the same as that of natural language, both with sequential structures. We call schedule primitives the tensor language. Thus, TLP is a \textbf{T}ensor \textbf{L}anguage \textbf{P}rocessing task that transforms the task of predicting tensor program latency with a cost model into an NLP regression task.

\subsection{Model Architecture}
\begin{figure}[t]%
    \centering
    
    \includegraphics[scale=0.7]{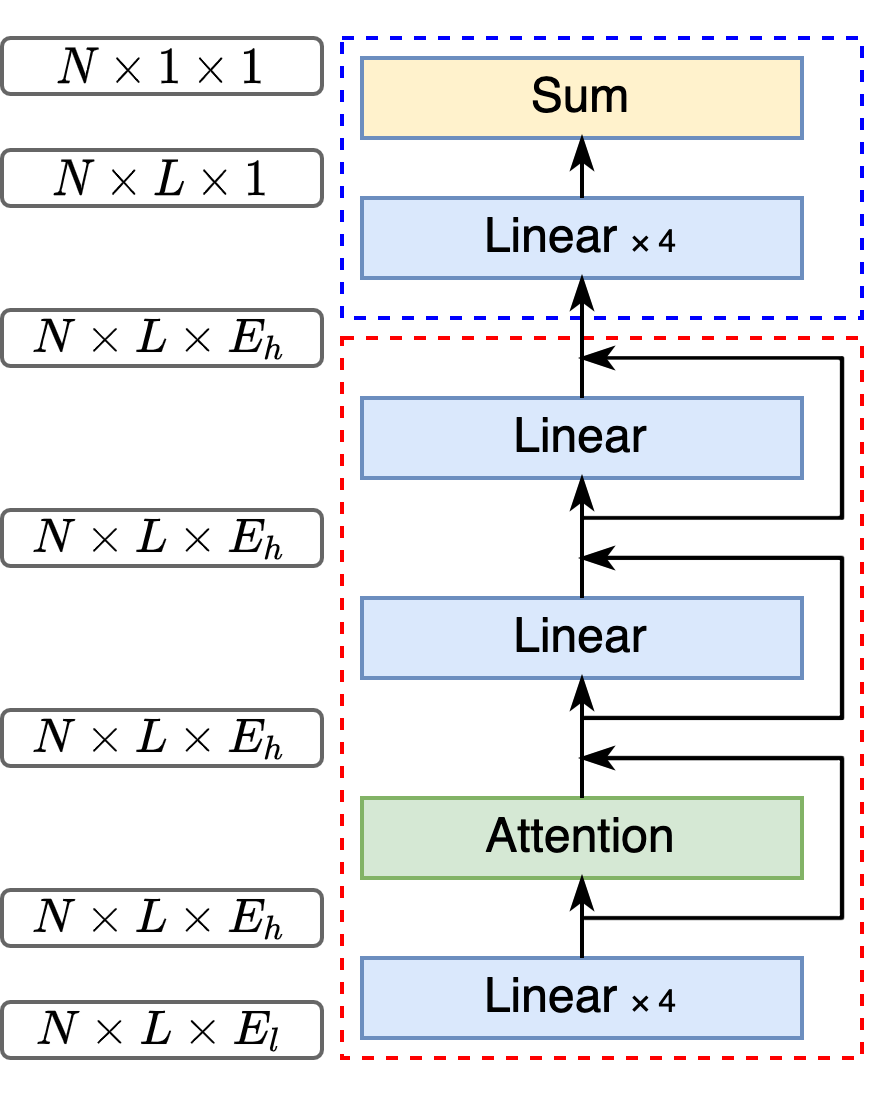}  

    \caption{The model architecture of TLP. The parameters on the left are the layer's input shape and output shape, where $N$ refers to the batch size, $L$ refers to the sequence length, and $E_{l}$ and $E_{h}$ refer to the embedding sizes. We call the red dashed box backbone and the blue dashed box head. These will be used in the multi-task learning chapter.}%
    \label{3_3_model}%

\end{figure}
Figure \ref{3_3_model} is the model architecture of TLP. The model first upsamples the dimension to 256 or more through multiple linear layers. Schedule primitives have strong contextual relationships. The self-attention mechanism can effectively capture contextual features, avoid the defect of memory forgetting, and run in parallel. In addition, LSTM can also effectively capture contextual features but cannot be executed in parallel. Therefore, we use the self-attention or LSTM module, which we call the backbone basic module, to capture contextual features. Then two residual blocks follow. Finally, multiple linear layers and a sum operation are used to obtain a prediction score. The target label is the normalized latency. The formula is as follows: $label={min\_latency}/{latency}$,
where min\_latency refers to the minimum value among all tensor programs of a subgraph. In the TenSet dataset, the number of tensor programs for a subgraph varies from dozens to 4,000. The value range of the label is (0, 1]. We use the Mean Square Error (MSE) loss function or the rank loss~\cite{cao2007learning, wang2018lambdaloss} function to calculate the loss.

\section{MTL-TLP}


\begin{table}[t]
\scriptsize
\centering
\begin{tabular}{|c|c|}
\hline
Cost Model & Representative Works \\ \hline
\begin{tabular}[c]{@{}c@{}}Empirical Formula\\ Cost Model\end{tabular} & Halide16~\cite{mullapudi2016automatically} \\ \hline
\begin{tabular}[c]{@{}c@{}}Online Learning\\ Cost Model\end{tabular} & \begin{tabular}[c]{@{}c@{}}AutoTVM~\cite{chen2018learning}, Ansor~\cite{zheng2020ansor}, Chameleon~\cite{ahn2020chameleon},\\ FlexTensor~\cite{zheng2020flextensor}, Halide20~\cite{adams2019learning, anderson2020learning}\end{tabular} \\ \cline{1-2}
\begin{tabular}[c]{@{}c@{}}Offline Learning\\ Cost Model\end{tabular} & \begin{tabular}[c]{@{}c@{}}TenSet, TIRAMISU cost model,\\ The work of Benoit Steiner et al.~\cite{steiner2021value}\end{tabular} \\ \hline
\end{tabular}
\caption{The development of tensor program cost models.}
\label{4_1_cost_models}
\end{table}
Table \ref{4_1_cost_models} shows the development of tensor program cost models. Early on, researchers directly used empirical formulas to evaluate the performance of tensor programs. For example, Halide16 used the data reuse rate and floating-point computation amount to evaluate the pros and cons of different loop-blocking and stage-folding strategies. Then, researchers began using simple deep learning models such as MLP, machine learning models such as XGBoost, reinforcement learning models, etc. These models are characterized by simple structures and low computational costs. In recent years, researchers have begun experimenting with complex deep learning models, such as TIRAMISU using the dynamic computational flow LSTM model and the work of Benoit Steiner et al. using the complex bilateral LSTM model. The law of this evolution is that the amount of data required has grown significantly simultaneously as performance has grown significantly. The online cost model only uses the measurement data generated during the tuning process and does not use any offline datasets. Offline cost models rely on massive offline datasets. The performance of the offline cost model is highly correlated with the size of the dataset. 

\subsection{Cross-hardware Unavailability}
The data collected offline are invalid across hardware caused by domain gaps in hardware architecture and hardware performance. This leads to a significant decrease in the performance of offline-trained cost models across hardware platforms. The empirical formula cost model and the online learning cost model do not suffer from this problem because they do not use offline datasets. 

This presents a challenge: how to use a small amount of target hardware data to train a cost model with high performance. The first thing that comes to mind when using a small amount of labeled data to obtain a high-performance model is the self-supervised learning method such as BERT~\cite{devlin2018bert, jiao2019tinybert} and GPT~\cite{brown2020language, radford2018improving, radford2019language}. However, it is worth noting that BERT generally uses 6-12 transformers layers, and the input data dimension is 512x768, etc. Nevertheless, our feature size is much smaller than that, not an order of magnitude with BERT's number of weights. For example, as mentioned earlier, the extracted features in TenSet are only 25x22. Such a large number of weights not only easily overfits the model but also severely slows down tuning, which is unacceptable during compilation. Another way to obtain a high-performance model with a small amount of data is transfer learning. There are many types of transfer learning, such as fine-tuning, training a local model to predict the gap between two domains, multi-task learning.

We think that multi-task learning techniques are most suitable for solving the problem of offline models being unavailable across hardware, mainly for the following three reasons:
\begin{itemize}
    \item The cross-hardware unavailability problem fits the scope of multi-task learning nicely, mitigating model performance degradation caused by domain gaps.
    \item Multi-task learning and the tensor compiler's multi-layer IRs share common design philosophy. The Tensor Compiler uses graph IR and tensor IR to abstract the hardware-independent and hardware-dependent features, respectively. Multi-task learning is a nice fit here: it can use shared parameters to fit hardware-independent features and non-shared parameters to fit hardware-dependent features.
    \item Our validation results also demonstrate that multi-task learning performs best among fine-tuning, multi-task learning, training a local model, and self-supervised learning. We will present these comparative data in the experimental section.
\end{itemize}

\subsection{MTL-TLP}
\begin{figure}[t]%
    \centering
    
    \includegraphics[scale=0.6]{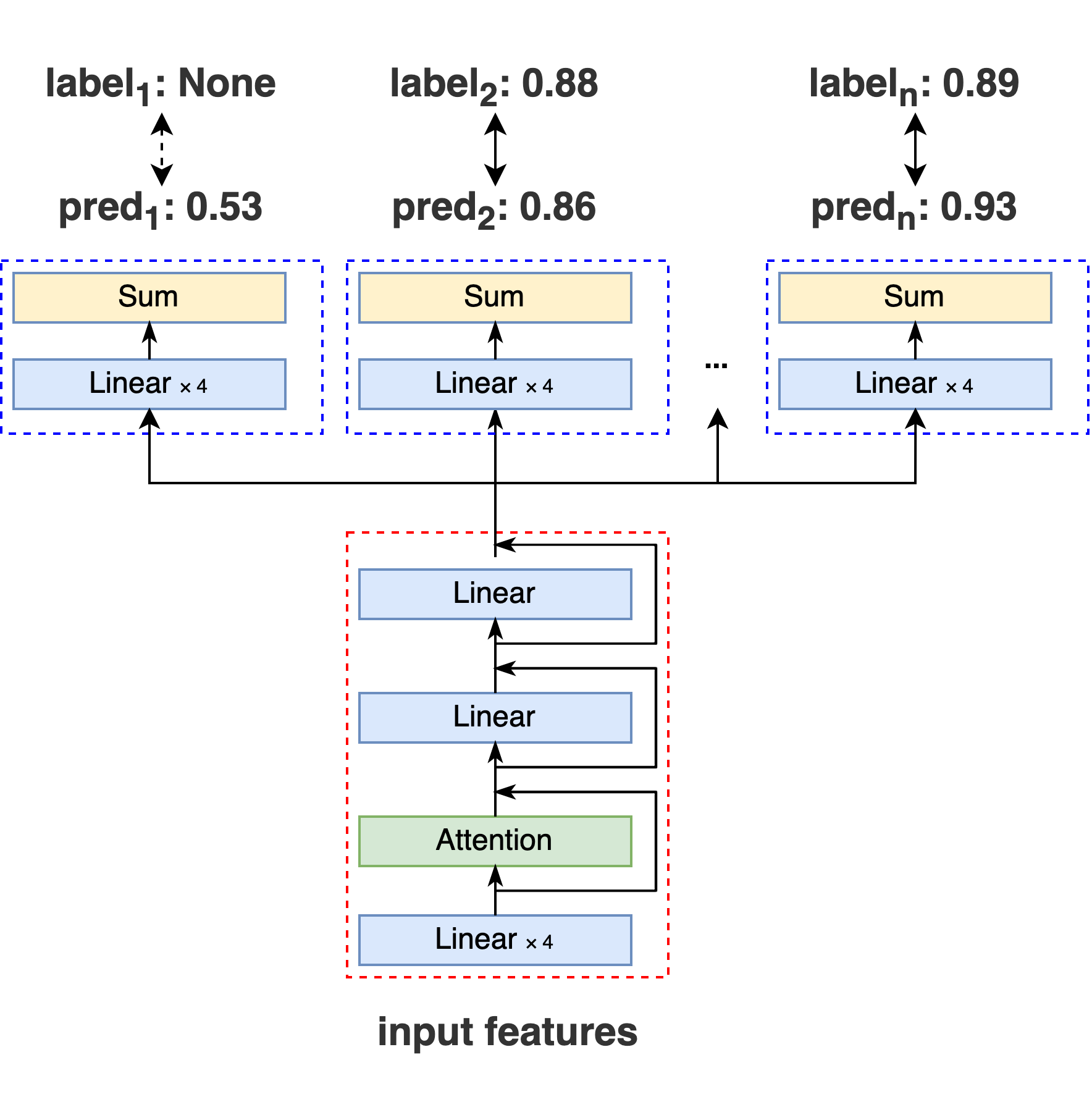}  

    \caption{The model architecture of MTL-TLP.}%
    \label{4_1_mlt_model}%

\end{figure}

The model structure of MTL-TLP is similar to that of TLP. The difference is that MTL-TLP sets up multiple heads, that is, multiple tasks, and each task corresponds to a hardware platform, as shown in Figure \ref{4_1_mlt_model}. If a labeled data of TLP is represented as a tuple such as $(input\ features,\ label)$, then the corresponding tuple of MTL-TLP is $(input\ features,\ [label_{1},\ label_{2},\ ...,\ label_{n}])$. We set task 1 as the task of the target platform. When the data has no label on the target platform, the tuple is $(input\ features,\ [None,\ label_{2},\ ...,\ label_{n}])$, and this target task will be ignored when calculating the loss, and the weights of this head will not be updated during back-propagation. The loss function is designed as follows:
\begin{center}
\begin{equation}
loss= \sum_{i=1}^{n}loss\_fun_{i}(pred_{i},\ label_{i}),\
where\ label_{i}\ is\ not\ None.\nonumber
\end{equation}
\end{center}
where $loss\_fun_{i}$ is the loss function of each task (usually the same), such as the MSE loss function, rank loss function, etc.

The tensor program is not universal between CPUs and GPUs, so we do not discuss multi-task learning across them.

\subsection{Feasibility Analysis of MTL-TLP}
MTL-TLP plays a big role mainly due to the following points:
\begin{itemize}
    \item MTL effectively increases the sample size that we are using for training our model. Although there is only a small amount of labeled data on the target platform, there is indeed a large amount of data on other platforms. The target platform cost model can be trained using data from other platforms as auxiliary tasks.
    \item In single-task learning, back-propagation of gradients tends to get stuck in local minima. In multi-task learning, the local minima of different tasks are in different positions. The interaction of different tasks can help the hidden layer escape from the local minima. 
    \item MTL acts as a regularizer by introducing an inductive bias. Multiple tasks share weights in shallow layers, which may weaken the network's ability, reduce network overfitting, and improve generalization.
    \item MTL-TLP can use shared parameters to fit hardware-independent features and non-shared parameters to fit hardware-dependent features.
\end{itemize}

\section{Evaluation}
There are two types of evaluation metrics to evaluate and compare the performance of cost models: dataset-based and search-based. The dataset-based metrics evaluate the model's accuracy on a static dataset. Search-based metrics evaluate the end-to-end search efficiency or search quality after integrating the cost models into the search algorithms.

\subsection{TLP with Dataset-based Metrics.}

We use data from the TenSet and TenSet-MLP datasets. At each hardware platform, we hold out a test set that consists of five networks, ResNet-50, MobileNet-V2, ResNext-50, BERT-tiny, and BERT-base, with batch size 1 and image size 224 (or sequence length 128). Divide all the remaining data into training and validation sets at a ratio of 9:1. Use the top-1 score and top-5 score as evaluation criteria, which have been proven effective in TenSet. The expression of top-k is as follows:
\begin{center}
\begin{equation}
top-k =  \frac{\sum_{m}{\sum_{s} {min\_latency_{m,s}\times weight_{m,s}}} }{\sum_{m}{\sum_{s}{\min_{} \left (latency_{m,s,i}\right)\times weight_{m,s}}} }   ,  1\le i\le k \nonumber
\end{equation}
\end{center}
where $min\_latency_{m,s}$ is the minimum latency among all tensor programs of subgraph $s$ of model $m$, $weight_{m,s}$ is the number of times the subgraph $s$ appears in model $m$, and $latency_{m,s,i}$ is the latency corresponding to the $i$-th largest value of the output score of the cost model for all tensor programs of subgraph $s$ of model $m$. If not explicitly stated, all top-1 scores in the following text refer to the top-1 scores on the five models in the above test set. The same goes for top-5.

\subsubsection{Loss Function \& Backbone Basic Module.}
First, we experiment with the loss function and the backbone basic module. For the loss function, the label of TLP is a normalization value in the range of (0,1], so MSE loss is an option. In addition, lambda rank loss designed for ranking tasks is also an option. For the backbone basic module, since the input feature of TLP is a sequence composed of schedule primitives, both LSTM and self-attention are suitable.

\begin{table}[b]
\scriptsize
\centering
\begin{tabular}{|c|c|c|}
\hline
 & Top-1 Score & Top-5 Score \\ \hline
Attention + Rank & \textbf{0.9194} & \textbf{0.9710} \\ \hline
Attention   + MSE & 0.9128 & 0.9542 \\ \hline
LSTM   + Rank & 0.9119 & 0.9509 \\ \hline
LSTM   + MSE & 0.9061 & 0.9540 \\ \hline
\end{tabular}
\caption{The top-k scores for different combinations of loss functions and backbone basic modules.}
\label{6_1_atention_mseloss}
\end{table}
We use the TenSet CPU dataset collected from Intel Xeon Platinum-8272. Table \ref{6_1_atention_mseloss} shows the top-k scores for different combinations of loss functions and backbone basic modules. We can find that self-attention + lambda rank loss is slightly more suitable for TLP than the other combinations. If not explicitly stated, all experiments in the following text use the combination of self-attention + lambda rank loss.

\subsubsection{Feature Size Cropping.}
As we discussed earlier, all schedule primitive sequences in the TenSet CPU dataset contain up to 54 schedule primitives, and each schedule primitive extracts up to 40 features. However, many are zero, leading to sparse features and affecting the cost model accuracy. We analyzed that reducing the feature size to 25$\times$22 is a good choice. We do experimental verification on this. 

\begin{table}[b]
\scriptsize
\centering
\begin{tabular}{|c|c|c|}
\hline
 & Top-1 Score & Top-5 Score \\ \hline
Seq Len 25 + Emb Size 22 & \textbf{0.9194} & \textbf{0.9710} \\ \hline
Seq Len 25 + Emb Size 40 & 0.9171 & 0.9558 \\ \hline
Seq Len 54 + Emb Size 22 & 0.9032 & 0.9472 \\ \hline
Seq Len 54 + Emb Size 40 & 0.9076 & 0.9677 \\ \hline
\end{tabular}
\caption{The top-k scores for different combinations of sequence lengths and embedding sizes.}
\label{6_2_transform_step_len}
\end{table}
We use the TenSet CPU dataset collected from Intel Xeon Platinum-8272. Table \ref{6_2_transform_step_len} shows the top-k scores for different combinations of sequence lengths and embedding sizes. The results show that reducing the sequence length and embedding size to 25 and 22, respectively, reduces the amount of computation and storage space while improving the accuracy of the cost model. We think this is because the features are denser at the cost of harming very few features, avoiding many zeros. If not explicitly stated, all experiments below use the combination of sequence length 25 + embedding size 22.

\subsubsection{Other Model Architecture Details.}
We also perform extensive experiments to determine some details of the model architecture. Here, we list the results. It is sufficient that shallow linear layers in the model upsample the embedding size from 22 to 256, and 512 or larger is not required. The self-attention module sets 8 heads with the highest accuracy. Using one layer of the self-attention module is enough, and there is no need to use multiple layers or transformers. The self-attention module followed by two residual blocks is more accurate than zero, one or more residual blocks. If not explicitly stated, these settings are used in subsequent experiments.

\subsubsection{Compare with TenSet MLP}
After the experiments above, the model architecture of TLP is determined. We compare it with the state-of-the-art TenSet MLP cost model. Compared with Ansor, TenSet MLP pre-trained on the TenSet dataset can accelerate the search time by up to 10$\times$. We use the TenSet and TenSet-TLP datasets collected from all platforms, including 5 CPUs and 2 GPUs.

\begin{table*}[t]
\scriptsize
\centering
\begin{tabular}{|c|cc|cc|}
\hline
 & \multicolumn{2}{c|}{TenSet} & \multicolumn{2}{c|}{Ours} \\ \hline
 & \multicolumn{1}{c|}{Top-1 Score} & Top-5 Score & \multicolumn{1}{c|}{Top-1 Score} & Top-5 Score \\ \hline
Intel Platinum 8272CL @ 2.60GHz (16   cores) & \multicolumn{1}{c|}{0.8748} & 0.9527 & \multicolumn{1}{c|}{\textbf{0.9194}} & \textbf{0.9710} \\ \hline
Intel E5-2673 v4 @ 2.30GHz (8 cores) & \multicolumn{1}{c|}{0.8332} & 0.8977 & \multicolumn{1}{c|}{\textbf{0.8941}} & \textbf{0.9633} \\ \hline
AMD EPYC 7452 @ 2.35GHz (4 cores) & \multicolumn{1}{c|}{0.8510} & 0.9175 & \multicolumn{1}{c|}{\textbf{0.9055}} & \textbf{0.9494} \\ \hline
ARM Graviton2 (16 cores) & \multicolumn{1}{c|}{0.7799} & 0.9049 & \multicolumn{1}{c|}{\textbf{0.8207}} & \textbf{0.9226} \\ \hline
Intel i7-10510U @ 1.80GHz (8 cores) & \multicolumn{1}{c|}{0.7776} & 0.8590 & \multicolumn{1}{c|}{\textbf{0.8473}} & \textbf{0.9427} \\ \hline
NVIDIA Tesla K80 & \multicolumn{1}{c|}{\textbf{0.9083}} & 0.9629 & \multicolumn{1}{c|}{0.9059} & \textbf{0.9741} \\ \hline
NVIDIA Tesla T4 & \multicolumn{1}{c|}{0.8757} & \textbf{0.9528} & \multicolumn{1}{c|}{\textbf{0.8847}} & 0.9250 \\ \hline
\end{tabular}
\caption{The top-k scores of TLP and TenSet MLP on all hardware platforms of the TenSet and TenSet-TLP datasets.}
\label{6_3_compare_wite_tenset}
\end{table*}
Table \ref{6_3_compare_wite_tenset} shows the top-k scores of TLP and TenSet MLP on all hardware platforms of the TenSet and TenSet-TLP datasets. It can be found that on all CPUs, TLP exceeds TenSet MLP by a large margin. On GPUs, TLP and TenSet MLP each have their strengths. However, later experiments prove that since TLP extracts features from schedule primitives rather than tensor programs, this saves much tuning time for TLP. TLP still outperforms TenSet MLP on GPUs. As seen from the top-5 scores, TLP has stable performance on different CPUs and GPUs. It is worth mentioning that TenSet MLP performs special feature extraction for CPUs and GPUs, and TLP uses the same mechanism to extract CPU and GPU features.

We can say that TLP feature extraction is a simple yet effective and general method. This is mainly due to a shift in our thinking about feature extraction. We do not extract features from the tensor program itself but schedule primitives. In the feature extraction process, we are not limited by prior knowledge and prune any features in advance because it cannot be proven that pruned features are detrimental to improving accuracy. We delegate this process to a neural network to weigh the importance of features through learnable weights.

\subsection{MTL-TLP with Dataset-based Metrics}
\subsubsection{Effectiveness of MTL-TLP}
To determine the effectiveness of MTL-TLP, we conduct experiments on CPUs and GPUs, respectively. We set up 4 experiments on CPUs, specifying Intel E5-2673 as the target hardware platform. The first experiment set up a task using 500K data collected from Intel E5-2673. The second experiment sets up two tasks: the first uses 500K data collected from Intel E5-2673, and the second uses all data collected from Intel Xeon Platinum-8272. The settings for the remaining two CPU experiments are similar. We set up 2 experiments on GPUs, specifying NVIDIA Tesla T4 as the target hardware platform. The experimental setup for GPUs is similar to that for CPUs.

\begin{table}[b]
{\scriptsize
\centering

\begin{tabular}{|c|c|c|c|}
\hline
 &  & Top-1 Score & Top-5 Score \\ \hline
E5-2673 & 500K & 0.6647 & 0.8848 \\ \hline
\begin{tabular}[c]{@{}c@{}}E5-2673\\      Platinum-8272\end{tabular} & \begin{tabular}[c]{@{}c@{}}500K\\      ALL\end{tabular} & 0.8741 & 0.9385 \\ \hline
\begin{tabular}[c]{@{}c@{}}E5-2673\\      Platinum-8272\\      EPYC-7452\end{tabular} & \begin{tabular}[c]{@{}c@{}}500K\\      ALL\\      ALL\end{tabular} & \textbf{0.8901} & \textbf{0.9520} \\ \hline
\begin{tabular}[c]{@{}c@{}}E5-2673\\      Platinum-8272\\      EPYC-7452\\      Graviton2\end{tabular} & \begin{tabular}[c]{@{}c@{}}500K\\      ALL\\      ALL\\      ALL\end{tabular} & 0.8753 & 0.9302 \\ \hline

\end{tabular}


\caption{The top-k scores on Intel E5-2673.}
\label{6_4_multi_task_cpu}
}
\end{table}

\begin{table}[b]
\scriptsize
\centering
\begin{tabular}{|c|c|c|c|}
\hline
 &  & Top-1 Score & Top-5 Score \\ \hline
Tesla   T4 & 500K & 0.7971 & 0.8984 \\ \hline
\begin{tabular}[c]{@{}c@{}}Tesla   T4\\      Tesla K80\end{tabular} & \begin{tabular}[c]{@{}c@{}}500K\\      ALL\end{tabular} & \textbf{0.8876} & \textbf{0.9373} \\ \hline
\end{tabular}
\caption{The top-k scores on NVIDIA Tesla T4.}
\label{6_4_multi_task_gpu}
\end{table}

Tables \ref{6_4_multi_task_cpu} and \ref{6_4_multi_task_gpu} show the top-k scores of multi-task learning for CPUs and GPUs, respectively. The top-k scores here are for the target hardware. The results show that using multi-task learning can significantly improve the accuracy of the cost model. Combining with Table \ref{6_3_compare_wite_tenset}, it can be found that even using only 500K data exceeds the TenSet MLP using all data and is on par with TLP using all data but without multi-task learning. Table \ref{6_4_multi_task_cpu} also shows that appropriately increasing the number of tasks can improve the accuracy of the cost model. However, if there are too many tasks, it will hinder the accuracy increase. This is because appropriately increasing the number of tasks can make the features learned by the MTL-TLP backbone more general and have a stronger generalization ability. However, if too many tasks exist, the interference between tasks increases, and sharing parameters cannot make the target platform task reach the optimal point.

\subsubsection{Transfer Learning \& Self-supervised Learning.}
\begin{table}[t]
\scriptsize
\centering
\begin{tabular}{|c|c|c|c|c|}
\hline
                             &                             &      & Top-1             & Top-5                       \\ \hline
\multirow{2}{*}{Fine-tuning} & E5-2673 (Pre-trained)       & ALL  & \multirow{2}{*}{0.7897} & \multirow{2}{*}{0.9175}          \\
                             & i7 (Fine-tuning)            & 500K &                         &                                  \\ \hline
\multirow{2}{*}{MTL}         & E5-2673 (Task 2)            & ALL  & \multirow{2}{*}{\textbf{0.8331}} & \multirow{2}{*}{\textbf{0.9672}} \\
                             & i7 (Task 1)                 & 500K &                         &                                  \\ \hline
\multirow{2}{*}{GPT}         & i7 (Pre-trained, Unlabeled) & ALL  & \multirow{2}{*}{0.6863} & \multirow{2}{*}{0.8431}          \\
                             & i7 (Downstream Task)        & 500K &                         &                                  \\ \hline
\multirow{2}{*}{BERT}        & i7 (Pre-trained, Unlabeled) & ALL  & \multirow{2}{*}{0.6316} & \multirow{2}{*}{0.8137}          \\
                             & i7 (Downstream Task)        & 500K &                         &                                  \\ \hline
\end{tabular}
\caption{The top-k scores for various transfer learning and self-supervised learning methods.}
\label{6_1_transfer_and_self_sup}
\end{table}
In this part, we will discuss various transfer learning and self-supervised learning methods. We set up 4 experiments: 1) Pre-training the cost model with all Intel E5-2673 data, then fine-tuning with 500K Intel i7-10510U data; 2) Training the cost model with multi-task learning techniques, task 1 uses 500K Intel i7-10510U data, task 2 uses all Intel E5-2673 data; 3) Use the cost model as a downstream task of GPT, first use all Intel i7-10510U data (without label) to pre-train a GPT model, and then use 500K Intel i7-10510U data training the cost model; 4) Similar to 3, but trains a BERT model. Table \ref{6_1_transfer_and_self_sup} shows the top-k scores for the above 4 experiments. Compared with fine-tuning, MTL is more suitable for solving the problem of unavailability across hardware. The two methods of GPT and BERT, as analyzed in the previous chapters, have a huge amount of parameters that are a burden for small-sized input features and are prone to overfitting.

\subsubsection{Multi-task Learning Between Architectures.}\label{MTL_arch}
We think that the effect of multi-task learning on the accuracy improvement of the target platform is different between different hardware architectures. To verify this, we set up four experiments with Intel i7-10510U as the target platform, and all experiments set up two tasks. All the first tasks use 500K data collected from Intel i7-10510U. The second task uses all data collected from Intel Platinum 8272, Intel E5-2673, AMD EPYC 7452, and ARM Graviton2, respectively.

\begin{table}[b]
\scriptsize
\centering
\begin{tabular}{|c|c|c|c|}
\hline
 &  & Top-1 Score & Top-5 Score \\ \hline
\begin{tabular}[c]{@{}c@{}}i7-10510U\\      Platinum-8272\end{tabular} & \begin{tabular}[c]{@{}c@{}}500K\\      ALL\end{tabular} & \textbf{0.8413} & 0.9202 \\ \hline
\begin{tabular}[c]{@{}c@{}}i7-10510U\\      E5-2673\end{tabular} & \begin{tabular}[c]{@{}c@{}}500K\\      ALL\end{tabular} & 0.8331 & \textbf{0.9672} \\ \hline
\begin{tabular}[c]{@{}c@{}}i7-10510U\\      EPYC-7452\end{tabular} & \begin{tabular}[c]{@{}c@{}}500K\\      ALL\end{tabular} & 0.8082 & 0.9122 \\ \hline
\begin{tabular}[c]{@{}c@{}}i7-10510U\\      Graviton2\end{tabular} & \begin{tabular}[c]{@{}c@{}}500K\\      ALL\end{tabular} & 0.7711 & 0.8909 \\ \hline
\end{tabular}
\caption{\small{The top-k scores between architectures.}}
\label{6_5_mtl_2head}
\end{table}
Table \ref{6_5_mtl_2head} shows the top-k scores for multi-task learning between architectures. The target hardware Intel i7-10510U is the Intel X86 architecture, and Intel Platinum 8272 and Intel E5-2673, both of which are X86, have the most considerable precision improvement on the target hardware. AMD and ARM CPUs also have improved precision on the target hardware, but they are not as good as the Platinum 8272 and Intel E5-2673.

\subsubsection{Multi-task Learning with Data Size.}\label{MTL_data_volume}
\begin{figure}[t]%
    \centering
    \includegraphics[scale=0.6]{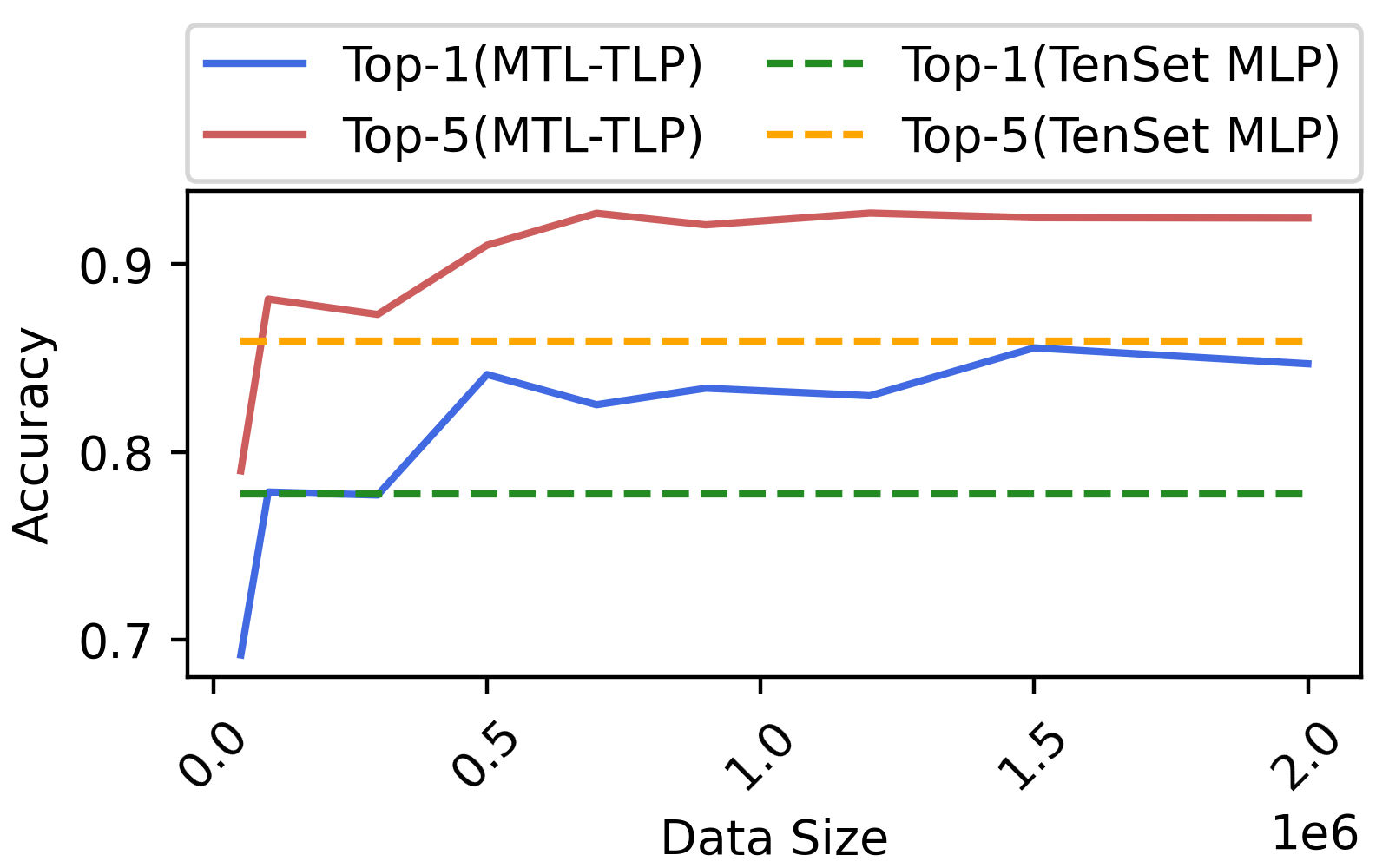}
    \caption{Accuracy curves of MTL-TLP using different data sizes.}%
    \label{5_2_3_mlt_curve}%

\end{figure}
We explore the relationship between MTL-TLP accuracy and target platform data size. The experiments set up two tasks, the second task uses all the data collected from Platinum 8272, and the first task uses 50K, 100K, 300K, 500K, 700K, 900K, 1.2M, 1.5M, and 2.0M data respectively. The results are shown in Figure \ref{5_2_3_mlt_curve}. When the data size increases to 500K, we see the most improvement, and the accuracy of MTL-TLP exceeds that of TenSet MLP.

Combining Sections \ref{MTL_arch} and \ref{MTL_data_volume}, when using MTL-TLP, we recommend setting up two to three tasks, with non-target platform tasks using all data from the same instruction set architecture hardware platform and the target platform task using at least 500K data.
\subsection{Search-based metrics}

End-to-end search efficiency and search quality are still the golden standards for checking the performance of a cost model. We integrate all the techniques discussed earlier into the auto-tuning framework Ansor and perform extensive experiments on both CPU and GPU.

The hardware platform of the CPU experiment is a notebook with an 8-core Intel i7-10510U, 16GB memory, and 2GB NVIDIA MX350. The hardware platform for GPU experiments is an 8-core Intel Platinum 8255C, 32GB memory, and 16GB NVIDIA Tesla T4. We perform the following four experiments on five models, ResNet-50, MobileNet-V2, ResNext-50, BERT-tiny, and BERT-base with batch size 1 and image size 224 (or sequence length 128): 1) Tuning with Ansor; 2) Tuning with TenSet MLP; 3) Tuning with TLP; 4) Tuning with MTL-TLP-500K. Among them, MTL-TLP-500K sets up two tasks. When training the model for the CPU platform, task one uses 500K Intel i7-10510U data, and task two uses all Intel Platinum-8272 data. When training the model for the GPU platform, task one uses 500K NVIDIA Tesla T4 data, and task two uses all NVIDIA Tesla K80 data.

TenSet transfer learning trains a local model to predict the gap between the source and target platforms. However, it does not rely on offline datasets of the target platform. For fairness, MTL-TLP trained with 500K target platform data is not directly compared with TenSet transfer learning but with TenSet MLP trained with all target platform data.

Here we briefly introduce Ansor's tuning process. Ansor will first generate some initial tensor programs for a subgraph according to predefined rules. Then use the cost model to pick out potential tensor programs. Use these potential tensor programs to generate more tensor programs through the genetic algorithm and use the cost model again to prune the poor performers. This step will iterate multiple times. Put the last selected $n$ tensor programs on the target machine to measure the latency. The above is called a tuning round. After Ansor tunes all subgraphs in a workload for one round, it divides the remaining time budgets into each subgraph according to the task schedule algorithm. All our experiments are tuned for 200 rounds, each round picking 10 tensor programs to measure, a total of 2,000 measurements, which we call tuning 2,000 times in the following.

\begin{figure}[t]%
    \centering
    \includegraphics[scale=0.65]{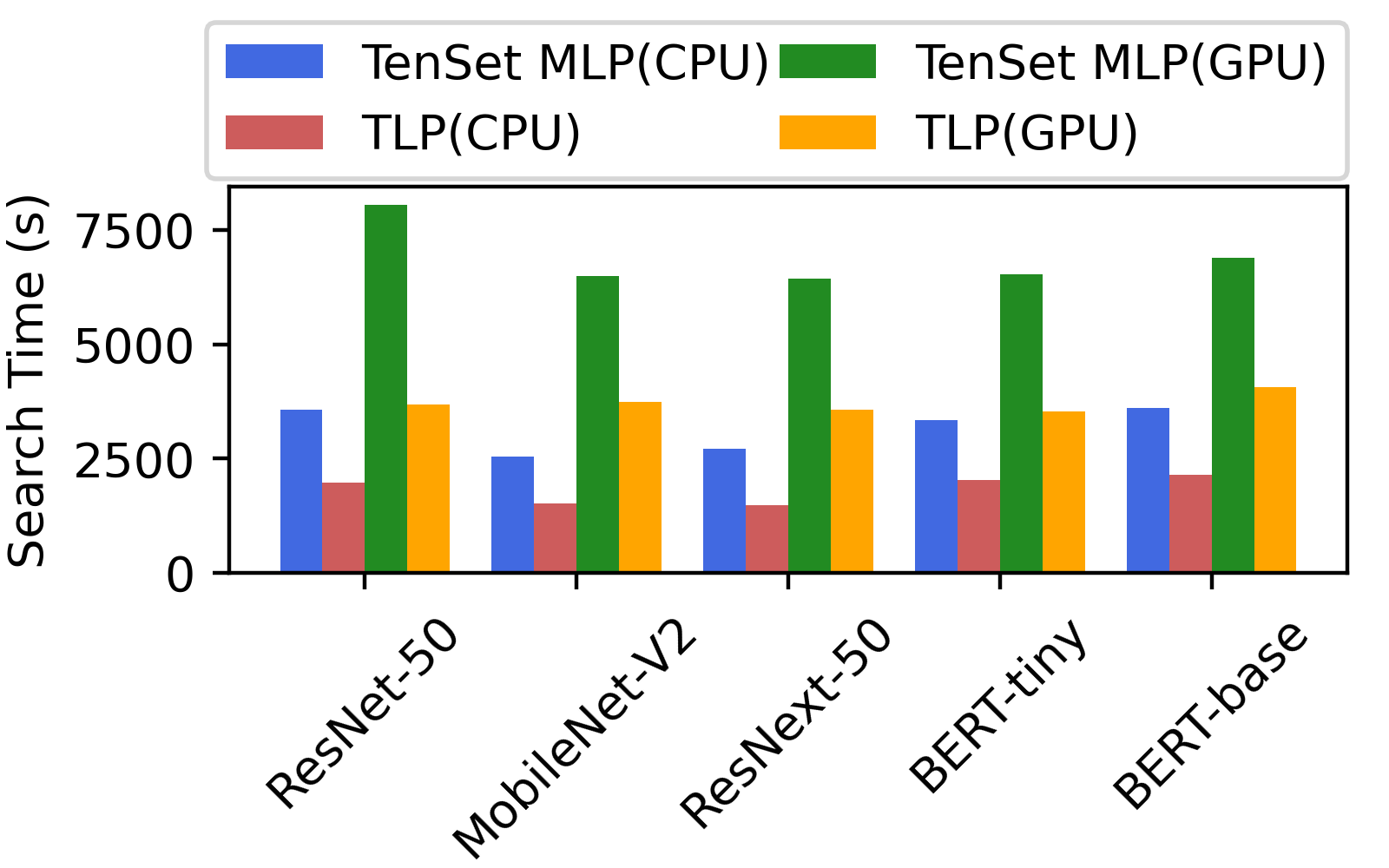} 
    \caption{Time in seconds for TLP and TenSet MLP to tune each model 2,000 times on the CPU and GPU.}%
    \label{6_3_run_times}%

\end{figure}
Figure \ref{6_3_run_times} shows the time, in seconds, taken by TLP and TenSet MLP to tune each model 2,000 times on the CPU and GPU. In terms of execution speed alone, TLP is on average 1.7$\times$ and 1.8$\times$ faster than TenSet MLP on the CPU and GPU, respectively, because TLP extracts features from schedule primitives without generating tensor programs.



\begin{figure*}[t]%
    \centering
    \includegraphics[scale=0.62]{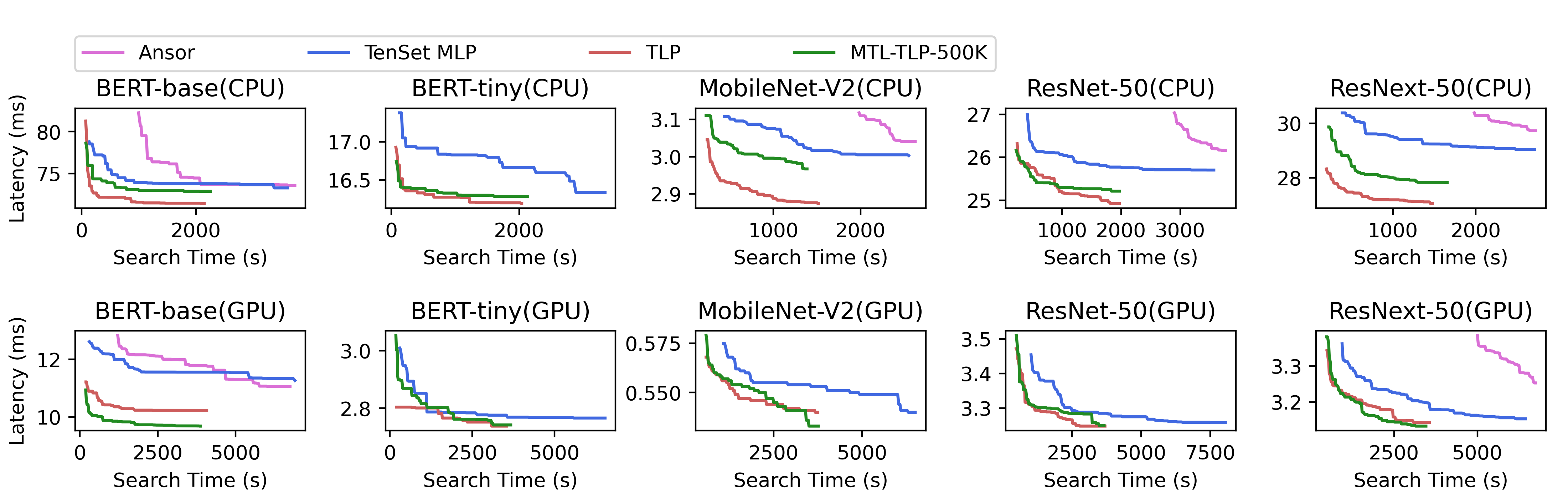}
    \caption{On the CPU and GPU, tuning curves for five workloads.}%
    \label{6_1_curve}%

\end{figure*}
\begin{figure}[t]%
    \centering
    \includegraphics[scale=0.62]{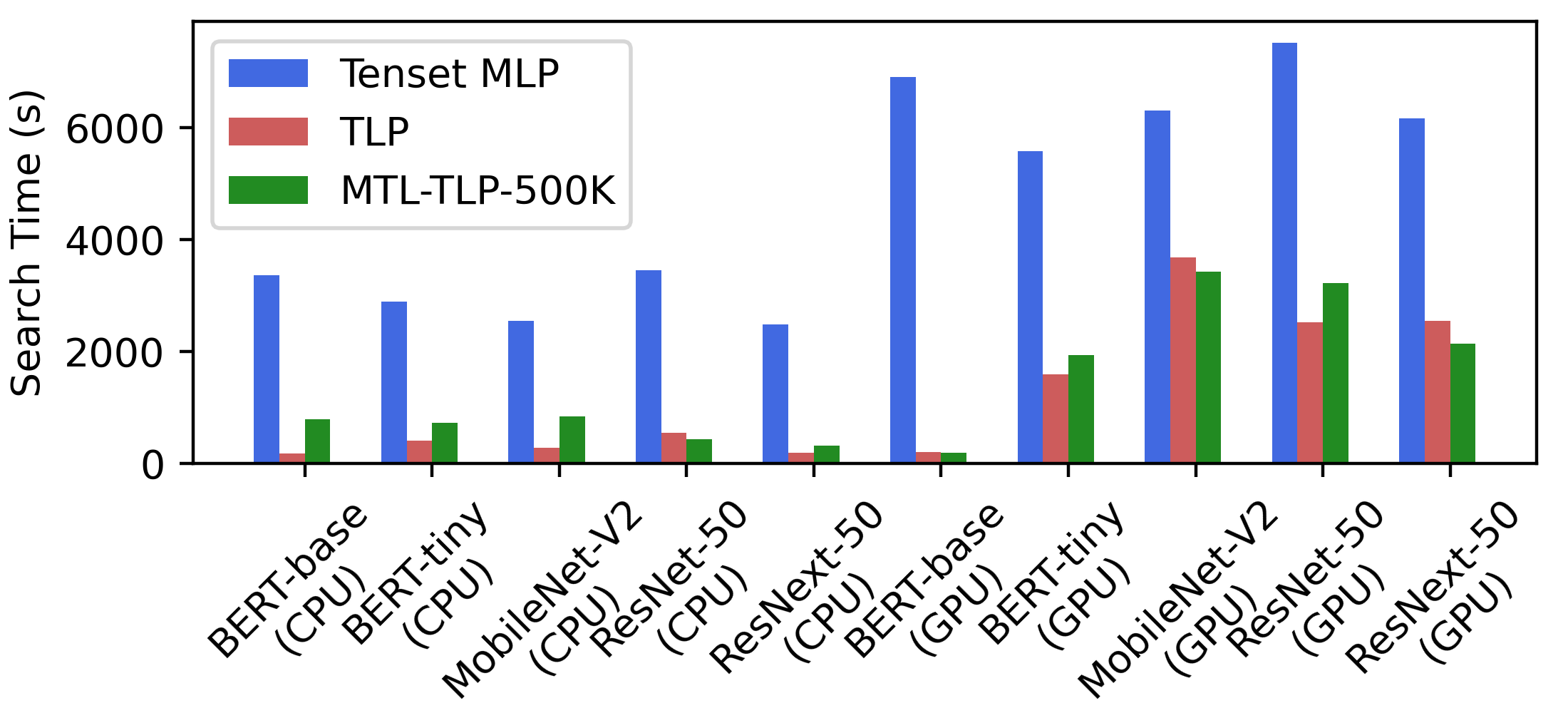} 
    \caption{On the CPU and GPU, the search time required for each deep learning workload with different cost models to reach the performance of TenSet MLP tuning 2,000 times.}%
    \label{6_2_search_time}%

\end{figure}
\begin{figure}[t]%
    \centering
    \includegraphics[scale=0.62]{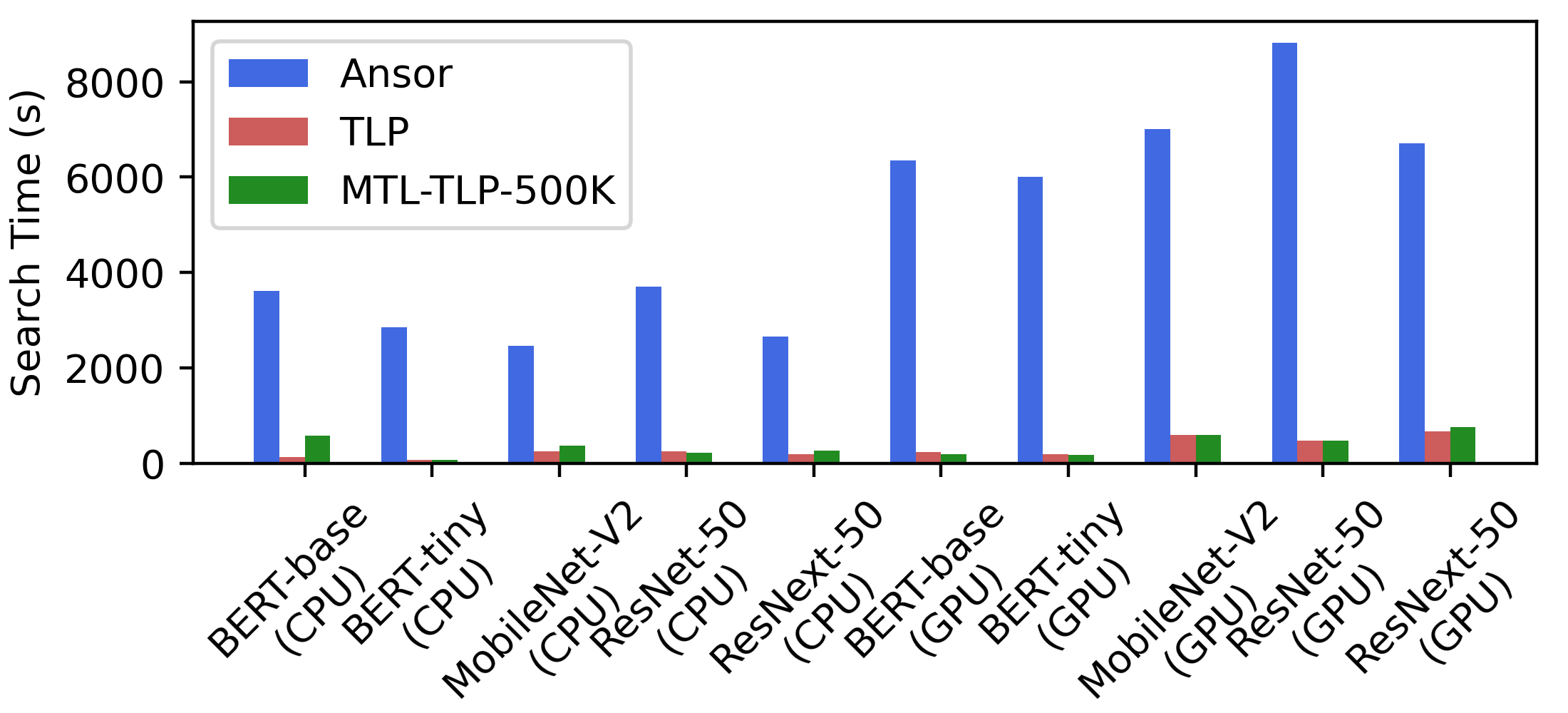} 
    \caption{On the CPU and GPU, the search time required for each deep learning workload with different cost models to reach the performance of Ansor tuning 2,000 times.}%
    \label{6_2_search_time_ansor}%

\end{figure}

Figure \ref{6_1_curve} shows the tuning curves for all workloads. To highlight the contrast between TLP, MTL-TLP, and TenSet MLP, the front part of Ansor's tuning curve is cropped. In some graphs, Ansor's curve disappears. This is because there is still a big gap between the performance of Ansor tuning 2000 times and the performance of TLP/MTL-TLP tuning the minimum times. The minimum times refers to the total times of tuning one round for each subgraph of the workload only. These curves show that TLP and MTL-TLP can converge to lower values faster. This trend is even more pronounced on CPUs. Figure \ref{6_2_search_time} shows the search time required for TLP, MTL-TLP, and TenSet MLP to achieve the performance of TenSet MLP tuning 2,000 times on the CPU and GPU. Compared with TenSet MLP, TLP can speed up the search time by an average of 9.1$\times$ and 3.0$\times$ on the CPU and GPU. MTL-TLP can speed up the search time by an average of 4.7$\times$ and 2.9$\times$ on the CPU and GPU. Figure \ref{6_2_search_time_ansor} shows the search time required to achieve the performance of Ansor tuning 2,000 times. Compared with Ansor, TLP can speed up the search time by an average of 16.7$\times$ and 16.0$\times$ on the CPU and GPU. MTL-TLP can speed up the search time by an average of 10.0$\times$ and 15.8$\times$ on the CPU and GPU. Among them, MTL-TLP only uses 500K data of the target platform, which is approximately 7\% of the TenSet dataset. As mentioned before, there is still a gap between Ansor tuning 2,000 times and TLP/MTL-TLP tuning only the minimum times for some workloads, so the speed-up compared to Ansor can be even larger.

\textbf{The Impact of TLP on Computing Resources.} 
In all the above end-to-end tuning experiments, approximately 10,000 schedule primitive sequences are performed for feature extraction and latency prediction for each subgraph in one round of tuning. The CPU performs feature extraction, and the GPU performs latency prediction. When batch\_size is set to 2048, compared to TenSet MLP, the GPU memory usage of TLP increases from 882MB to 1634MB. This is not a problem, for even the GPU devices of ordinary notebooks can be competent. The time to execute five rounds of the genetic algorithm is reduced from about 20 seconds to about 6 seconds, including the time for latency prediction. In short, the resource usage of TLP on the CPU decreases, and the resource usage on the GPU increases.

\section{Related Work} 
Recently, the rapid development of tensor compilers has spawned several schemes, including Halide~\cite{ragan2013halide}, TVM~\cite{chen2018tvm}, XLA~\cite{XLA}, nGraph~\cite{cyphers2018intel}, TIRAMISU~\cite{baghdadi2019tiramisu}, DeepCuts~\cite{jung2021deepcuts}, TACO~\cite{kjolstad2017tensor}, and Tensor Comprehensions~\cite{vasilache2018tensor}.

\textbf{Auto-tuners and cost models.}
AutoTVM~\cite{chen2018learning} is a template-based automatic search framework, the first search framework integrated in TVM. Subsequently, to deal with the limited search space caused by templates, Ansor~\cite{zheng2020ansor} proposed an automatic search framework based on hierarchy. Bolt~\cite{xing2022bolt} bridges the performance gap between tensor compilers and hardware-native libraries using hardware-native templated search enabled by modular and configurable libraries provided by hardware vendors. Benoit Steiner et al.~\cite{steiner2021value} use the value learning of reinforcement learning and a bidirectional LSTM cost model to solve the automatic tuning problem. It adopts a reinforcement learning-based search strategy, using the cost of the tensor program as the reward value during reinforcement learning. ProTuner~\cite{haj2020protuner} uses Monte Carlo tree search to solve the inaccurate estimation problem in the Halide auto-scheduler. The TIRAMISU cost model~\cite{baghdadi2021deep} extracts loop information, buffer access matrix, and AST information from the tensor program and propagates the data forward recursively according to the AST structure. FlexTensor~\cite{zheng2020flextensor} proposes a general template to cover multiple operators, and its template is designed for single operator granularity. 
The polyhedral compilation model~\cite{bondhugula2008practical, baghdadi2015pencil, verdoolaege2016presburger, verdoolaege2013polyhedral} formalizes the problem of automatic code optimization as an integer linear programming (ILP) problem. It optimizes a program with affine loop transformation that minimizes the data reuse distance between dependent statements. DeepTune~\cite{cummins2017end} proposes to predict whether an OpenCL kernel should be mapped to CPU or GPU.

\textbf{Tensor program datasets.}
TenSet~\cite{zheng2021tenset} is a large-scale tensor program dataset that continues Ansor's work. 
The TIRAMISU cost model and the work of Benoit Steiner et al. ~\cite{steiner2021value} also collect some tensor program datasets, but neither is open source.

\textbf{Cross-hardware solutions.}
There are few studies on cost models across hardware platforms. TenSet builds a local model to predict the latency gap between the source and target hardware. Moses~\cite{zhao2022moses} uses model distillation to distill out transferable and non-transferable parameters.

\section{Discussion}
\textbf{Limitation.}
Although we significantly address the unavailability of cost models across platforms, MTL-TLP trained with only 7\% of the target platform data outperforms TenSet MLP with all data. However, it still takes tens of hours to collect 500K data. The cross-hardware unavailability requires more research.

\noindent\textbf{Future Work.}
We have demonstrated the feasibility of schedule primitives as tensor program equivalents. 
We implement TLP and MTL-TLP in Ansor, which can be easily migrated to other tensor compiler automatic search frameworks. We have transformed the tensor program tuning into an NLP regression task. Based on this, it can be tried to use more mature NLP techniques to solve tensor program problems.
\section{Conclusion}
This paper proposes TLP and MTL-TLP. TLP designs a simple yet effective and general feature extraction mechanism for tensor program tuning. MTL-TLP utilizes multi-tasking techniques to solve the offline cost model cross-hardware unavailability. We analyze the feasibility and advantages of schedule primitives as objects for feature extraction. We theoretically analyze the reasons why MLT-TLP is effective. In the experimental part, we verify the effectiveness of TLP and MTL-TLP from database-based and search-based metrics. Experiments show that TLP and MTL-TLP outperform state-of-the-art implementations by a large margin. 
\section*{acknowledgements}
This work was supported in part by the National Key Research and Development Program of China under Grants No. 2018AAA0100500, National Natural Science Foundation of China under Grant No. 62272434, and Frontier Scientific Research Program of the Chinese Academy of Sciences  ZDBS-LY-JSC001.

\bibliographystyle{plain}
\bibliography{references}

\end{document}